\pdfoutput=1

\documentclass[11pt]{article}

\usepackage[]{acl}

\usepackage{times}
\usepackage{graphicx}
\usepackage{latexsym}
\usepackage{tcolorbox}
\usepackage{fancyvrb}
\usepackage{listings}
\usepackage{booktabs}
\usepackage{arydshln}

\usepackage[T1]{fontenc}

\usepackage[utf8]{inputenc}

\usepackage{microtype}

\title{Learning to Reason: Training LLMs with GPT-OSS or DeepSeek R1 Reasoning Traces}

\author{Shaltiel Shmidman\textsuperscript{1,†}, Asher Fredman\textsuperscript{2,‡}, Oleg Sudakov\textsuperscript{2,‡}, Meriem Bendris\textsuperscript{2,‡} \\
\textsuperscript{1}DICTA \quad
\textsuperscript{2}NVIDIA \\
\texttt{\small \textsuperscript{†}shaltiel@dicta.org.il} \\
\texttt{\small \textsuperscript{‡}\{afredman,osudakov,mbendris\}@nvidia.com}}

\begin{document}
\maketitle

\begin{abstract}

Test-time scaling, which leverages additional computation during inference to improve model accuracy, has enabled a new class of Large Language Models (LLMs) that are able to reason through complex problems by understanding the goal, turning this goal into a plan, working through intermediate steps, and checking their own work before answering . Frontier large language models with reasoning capabilities, such as DeepSeek-R1 and OpenAI’s gpt-oss, follow the same procedure when solving complex problems by generating intermediate reasoning traces before giving the final answer. Today, these models are being increasingly used to generate reasoning traces that serve as high-quality supervised data for post-training of small and medium-sized language models to teach reasoning capabilities without requiring expensive human curation.
In this work, we compare the performance of medium-sized LLMs on Math problems after post-training on two kinds of reasoning traces. We compare the impact of reasoning traces generated by DeepSeek-R1 and gpt-oss LLMs in terms of accuracy and inference efficiency.

\end{abstract}

\section{Introduction}

One way to improve the reasoning capabilities in LLMs is using test-time scaling, which leverages additional compute during inference to improve accuracy. This was originally accomplished via \textit{Chain-Of-Thought Prompting} \cite{wei2023chainofthoughtpromptingelicitsreasoning}, where the model was prompted to first reason through a problem before arriving at the solution. Then, reasoning models were developed such as \texttt{ChatGPT-o1} \cite{OpenAI2024o1SystemCard} and \texttt{Gemini-2.5} \cite{comanici2025gemini25pushingfrontier}, which "think out loud" by generating a sequence of reasoning tokens (reasoning trace) before providing an answer without explicitly exposing these traces to the user.  The release of open source large-scale reasoning models such as \texttt{DeepSeek-R1} \cite{deepseekai2025deepseekr1incentivizingreasoningcapability} and the OpenAI \texttt{gpt-oss} series \cite{OpenAI2025GPT‐OSS} has enabled the community to access the reasoning traces of these models. 

With the rise of frontier reasoning LLMs created with novel post-training techniques \cite{deepseekai2025deepseekr1incentivizingreasoningcapability,zheng2025groupsequencepolicyoptimization,lambert2025tulu3pushingfrontiers}, including verifier-guided RL, rule-based format/accuracy rewards, rejection sampling, and distillation, many model builders can now leverage synthetic data produced by large-scale reasoning models. For example, to enable the generation of reasoning traces by DeepSeek R1\cite{deepseekai2025deepseekr1incentivizingreasoningcapability}, a combination of human-annotated cold-start data, reasoning-oriented RL, custom loss function and specialized training techniques was used. This data serves as high-quality supervised data to fine-tune small/medium-sized models for reasoning.  
 
However, synthetic reasoning traces produced by these models vary significantly in both style and verbosity. This motivates the central research question: which reasoning style is more effective to distill into small/medium-sized models.

In this work, we compare reasoning styles of two open source LLMs: DeepSeek-R1 and gpt-oss. We measure their impact when fine-tuning 12B-parameter-sized base models on reasoning data, in terms of accuracy as well as inference efficiency. We focus on Math problems as they require complex problem-solving skills and are heavily affected by Test-Time Scaling.

\section{Math Reasoning Dataset}

In our experiments, 300,000 math conversations were sampled from the \texttt{Nemotron-Post-Training-Dataset-v1} \cite{bercovich2025llamanemotronefficientreasoningmodels,NemotronPostTrainingDatasetV1}. This dataset has a "math" split which provides math problems of various difficulty with their ground truth answers, as well as the corresponding generated responses from \texttt{DeepSeek-R1-2508} with full reasoning traces. 
To compare the reasoning styles between two models, we used \texttt{gpt-oss-120b} model to sample a response for each of the 300,000 questions\footnote{We did not use any system prompt when sampling, so the model uses the default reasoning effort.}, storing the full response with the reasoning traces.

Then we filtered out samples in which the answers generated by either \texttt{DeepSeek-R1-2508} or \texttt{gpt-oss-120b} didn't match the ground truth. To automate the filtering, \texttt{Qwen3-30B-A3B-Thinking-2507} \cite{qwen3technicalreport} was used as the judge model. The model was provided the ground truth answer and the generated answer, and prompted to determine whether the answers were the same. The full system prompt used can be found in Appendix \ref{sec:appendix_systemprompptp}. Finally, only samples where both models produced the correct answers were selected. The final dataset consists of 242,000 samples (each sample consists of a math problem, ground truth answer, and full reasoning trace from both DeepSeek-R1 and gpt-oss). 

Analyzing the reasoning trace styles, we observed that \texttt{DeepSeek-R1} generated on average $4.4\times$ as many tokens compared to \texttt{gpt-oss-120b}:
\begin{itemize}
    \item \textbf{\texttt{DeepSeek-R1}} response length was on average $\approx 15,500$ tokens.
    \item \textbf{\texttt{gpt-oss-120b}} response length was on average $\approx 3,500$ tokens.
\end{itemize}

\section{Experimental Setup}

In our experiment, we trained two LLMs on the two distinct reasoning styles, and evaluated their performance on various Math benchmarks. 

\subsection{Base Models}

We conducted out experiments with two 12B parameter models:

\begin{itemize}

    \item \textbf{\texttt{{\fontsize{10.2pt}{11pt}\selectfont NVIDIA-Nemotron-Nano-12B-v2-Base}}} \cite{nvidia2025nvidianemotronnano2} - a high-performing 12B base model pretrained from scratch by NVIDIA, which was infused with reasoning traces from \texttt{DeepSeek-R1} during the mid-training. 
    
    \item \textbf{\texttt{{\fontsize{10.2pt}{11pt}\selectfont Mistral-Nemo-Base-2407}}} \cite{MistralNemoBase2407} - a high-performing 12B model. It was chosen for comparison due to its release prior to reasoning models, and therefore not having any reasoning traces in the pretraining dataset.

\end{itemize}

\subsection{Infrastructure}

We ran our experiments on a cluster of H200 141GB GPUs on NVIDIA DGX Cloud Lepton\cite{anand_soman_2025_dgx_cloud_lepton}. All training was done using NVIDIA NeMo Framework \cite{Harper_NeMo_a_toolkit}, a scalable generative AI framework built for researchers and developers working on large language models, optimized for large-scale model training on NVIDIA hardware.

Models evaluation was performed using NVIDIA NeMo-Skills\footnote{\url{https://github.com/NVIDIA-NeMo/Skills/}}, following the same settings described in \citet{nvidia2025nvidianemotronnano2}.

\subsection{Training Details}

Each training session ran for 3,000 steps, with a global batch size of 64 ($\approx 4M$ tokens / step, $\approx 11.5B$ tokens total). We used a learning rate of $5e-6$, with a warmup ratio of $0.03$ from $5e-7$. We used the \texttt{AdamW}\cite{loshchilov2019decoupledweightdecayregularization} optimizer with a cosine annealing schedule \cite{loshchilov2017sgdrstochasticgradientdescent}.
Each sample was compiled using the \texttt{NVIDIA-Nemotron-Nano-12B-v2} chat template\footnote{\url{https://huggingface.co/nvidia/NVIDIA-Nemotron-Nano-12B-v2/blob/main/tokenizer_config.json\#L8008}}. All samples were then packed into training samples of 60k tokens, using the \textit{first fit decreasing} algorithm. Following the community standard,  we only computed the loss on the completions during training. 
The full example code for training is available\footnote{\url{https://gist.github.com/shaltielshmid/af27fc1ac24fcb85592bbf12dadcd10f}}.

\section{Results}
\begin{table*}[ht]
\centering
\renewcommand{\arraystretch}{1.15}
\begin{tabular}{@{}l l r r : r r@{}}
\toprule
 &  & \multicolumn{2}{c}{\textbf{Pass@8}} & \multicolumn{2}{c}{\textbf{Avg Tokens Generated}} \\
\textbf{} & \textbf{Benchmark} & \textbf{gpt-oss} & \textbf{DeepSeek-R1} & \textbf{gpt-oss} & \textbf{DeepSeek-R1} \\
\midrule
\textbf{Mistral-Nemo-Base-2407}& AIME25   & 30.0 & 23.3 & 12{,}951 & 29{,}328 \\
                 & GSM8K    & 96.4& 97.2 & 609 & 3{,}883 \\
                 & MATH-500 & 88.8 & 88.0 & 3{,}456 & 14{,}293 \\
\midrule\textbf{Nemotron-Nano-12B-v2-Base}& AIME25   & 83.3 & 83.3 & 7{,}980 & 20{,}179 \\
                 & GSM8K    & 97.0 & 97.7 & 375 & 2{,}010 \\
                 & MATH-500 & 98.0 & 99.0 & 1{,}350 & 5{,}931 \\
\bottomrule
\end{tabular}
\caption{Pass@8 accuracy and average token usage for training Mistral-Nemo 12B and Nano-V2-12B on the two datasets - gpt-oss vs. DeepSeek-R1. Both datasets produce similar accuracy, while the model trained using \texttt{DeepSeek-R1} traces produces $4\times$ as many tokens on average.}
\label{tab:results}
\end{table*}

\begin{figure}
  \centering
  \includegraphics[width=\linewidth]{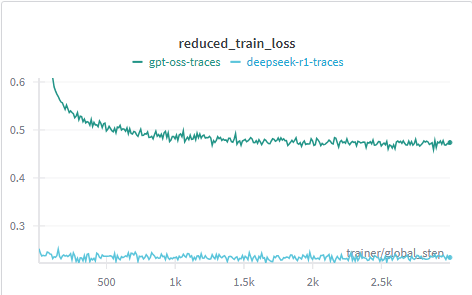}
  \caption{Graph comparing the training loss when fine-tuning \texttt{Nemotron-Nano-12B-V2} on the two datasets. The \texttt{DeekSeek-R1} dataset doesn't seem to significantly affect the loss, which is most likely because of the introduction of the reasoning traces in the mid-training.}
  \label{fig:loss-graph}
\end{figure}

We evaluated the trained models on three popular math benchmarks: GSM8k \cite{cobbe2021trainingverifierssolvemath}, AIME 2025 \cite{aime25_benchmark}, and MATH-500 \cite{math500_benchmark}. All models were evaluated under the same conditions - temperature $=0.6$, top\_p $=0.95$, tokens\_to\_generate $=32768$, and number\_of\_repeats$=8$.

Table \ref{tab:results} shows the results of the trained models using both DeepSeek-R1 and gpt-oss reasoning styles, evaluated on Math benchmarks. We observed that both tracing styles deliver similar downstream accuracy, with the \texttt{gpt-oss} style generating $4\times$ fewer tokens per response on average.

Figure \ref{fig:loss-graph} shows the training loss when fine-tuning \texttt{Nemotron-Nano-12B-V2} on the two datasets. We observed that the fine-tuning loss on traces generated by \texttt{DeepSeek-R1} starts very low and stays more or less constant throughout the training. In contrast, the fine-tuning loss on reasoning traces generated by gpt-oss, starts much higher and decreases gradually.  This is explained by the fact that the \texttt{Nemotron-Nano-12B-v2} mid-training dataset already included reasoning traces generated from \texttt{DeepSeek-R1}.

\section{Discussion}

As we saw in Table \ref{tab:results}, the \texttt{gpt-oss} reasoning style achieves accuracy comparable to the \texttt{DeepSeek-R1} style, while requiring significantly fewer tokens. This preliminary comparative analysis indicates that reasoning with more tokens does not systematically lead to better performance and that a trade-off can be be found between the number of tokens and model accuracy when solving Math problems.

Moreover, our experiments show that even when a reasoning model, such as Nemotron-Nano-12B-V2, has been trained on the more verbose reasoning style, it is possible to teach it to produce correct answers with significantly fewer tokens. This indicates that models are likely not permanently locked into a single reasoning format and can be effectively trained to adopt a new, more efficient one.

The most significant implication of this finding is related to inference efficiency. Models trained on the \texttt{gpt-oss} traces generated, on average, $4\times$ fewer tokens during evaluation. In real-world applications, this translates directly to a $\approx 4\times$ reduction in latency and cost for generating responses. For applications built on test-time scaling, this efficiency gain is non-negligible, enabling faster user experiences and substantially lower operational costs.

For future work, expanding this study to other domains and larger model scales would be highly valuable. 
Indeed, while math is an excellent proxy for complex, multi-step reasoning, our observations may not generalize to other domains like coding, creative writing, or general instruction-following, where verbosity might play a different role. In addition, we would like to explore a hybrid training approach, mixing both reasoning styles during training and investigate whether models can learn to apply the optimal level of verbosity for a given problem's difficulty.

\section{Conclusion}

In this work, we compare the impact of training medium-sized language models to utilize test-time scaling with two different reasoning styles generated from DeepSeek-R1 and gpt-oss. Our experiments demonstrate that both styles achieve comparable accuracy on math benchmarks, while exhibiting different degrees of verbosity. The findings here suggest that verbose reasoning on a higher number of tokens does not necessarily translate to better performance, and that models can be trained to adopt more efficient reasoning patterns. The reduction in the number of generated tokens directly affects deployment costs and latency in production systems. We hope this work encourages further investigation into the relationship between reasoning trace characteristics and both model performance and inference efficiency. Dicta is making the dataset from this research available on HuggingFace\footnote{\url{https://huggingface.co/datasets/dicta-il/MathCOT-oss-vs-DeepSeek}} to support further research and serve as a resource for the community.

\section{Acknowledgments}

We would like to thank the NVIDIA DGX Cloud Lepton team for early access to the platform, and for providing us with the necessary compute to train this model. 

\bibliography{anthology}
\bibliographystyle{acl_natbib}

\newpage

\appendix

\onecolumn

\section{Appendix: System Prompt used for verifying correctly generated answer}
\label{sec:appendix_systemprompptp}

\begin{tcolorbox}[colback=gray!5!white, colframe=gray!80!black,
  sharp corners, boxrule=0.7pt, 
  top=2mm, bottom=2mm, left=2mm, right=2mm]

\begin{lstlisting}[basicstyle=\ttfamily\small, breaklines=true, columns=fullflexible, keepspaces=true]
You are a meticulous and intelligent judge of mathematical solutions. Your sole purpose is to compare a <GENERATED_ANSWER>, which provides a full step-by-step trace to a problem, with a <CORRECT_ANSWER>, which contains only the final, verified solution.

Your task is to determine if the final conclusion of the <GENERATED_ANSWER> is mathematically equivalent to the <CORRECT_ANSWER>.

**Instructions:**

1.  **Identify the Final Answer:** Carefully parse the <GENERATED_ANSWER> to locate its final conclusion. This may be at the end of the text, often after a phrase like "Therefore, the answer is" or enclosed in a LaTeX \boxed{}.
2.  **Compare for Mathematical Equivalence:** Compare the extracted final answer from <GENERATED_ANSWER> with the content of <CORRECT_ANSWER>.
3.  **Handle Discrepancies:**
    *   **LaTeX Formatting:** Do not be concerned with differences in LaTeX syntax if the rendered mathematical expression is identical. For example, x=5 is the same as x = 5.
    *   **Multiple Solutions:** The <GENERATED_ANSWER> may offer more than one possible solution. The check should pass if the <CORRECT_ANSWER> is present as one of these solutions.
4.  **Provide a Verdict:**
    *   If the final answers are mathematically equivalent, respond with **MATCH**.
    *   If they are not equivalent, respond with **MISMATCH**.

Think very hard and make sure that you never produce a **MATCH** when it's not correct. Analyze the inputs and provide your verdict.
\end{lstlisting}

\end{tcolorbox}

\end{document}